\documentclass[a4paper,twoside]{article}

\usepackage{epsfig}
\usepackage{subcaption}
\usepackage{calc}
\usepackage{amssymb}
\usepackage{amstext}
\usepackage{amsmath}
\usepackage{amsthm}
\usepackage{multicol}
\usepackage{pslatex}
\usepackage{apalike}
\usepackage[bottom]{footmisc}
\usepackage{multirow}

\usepackage{mathtools}
\usepackage{hyperref}
\hypersetup{
    colorlinks=true,
    linkcolor=blue,
    filecolor=magenta,      
    urlcolor=cyan,
    pdftitle={VISAPP 2022 Mridula Painting},
    pdfpagemode=FullScreen,
    }

\usepackage{SCITEPRESS}     

\usepackage{xcolor}

\begin{document}


\title{Tackling Data Bias in Painting Classification with Style Transfer}

\author{\authorname{Mridula Vijendran\orcidAuthor{0000-0002-4970-7723}, Frederick W. B. Li\orcidAuthor{0000-0002-4283-4228} and Hubert P. H. Shum\orcidAuthor{0000-0001-5651-6039}\thanks{Corresponding author.}}
 \affiliation{Department of Computer Science, Durham University, Durham, UK}
 \email{\{mridula.vijendran, frederick.li, hubert.shum\}@durham.ac.uk}
}

\keywords{Data bias, style transfer, image classification, deep learning, paintings.}

\abstract{It is difficult to train classifiers on paintings collections due to model bias from domain gaps and data bias from the uneven distribution of artistic styles. Previous techniques like data distillation, traditional data augmentation and style transfer improve classifier training using task specific training datasets or domain adaptation. We propose a system to handle data bias in small paintings datasets like the Kaokore dataset while simultaneously accounting for domain adaptation in fine-tuning a model trained on real world images. Our system consists of two stages which are style transfer and classification. In the style transfer stage, we generate the stylized training samples per class with uniformly sampled content and style images and train the style transformation network per domain. In the classification stage, we can interpret the effectiveness of the style and content layers at the attention layers when training on the original training dataset and the stylized images. We can tradeoff the model performance and convergence by dynamically varying the proportion of augmented samples in the majority and minority classes. We achieve comparable results to the SOTA with fewer training epochs and a classifier with fewer training parameters.}

\onecolumn \maketitle \normalsize \setcounter{footnote}{0} \vfill


\section{\uppercase{Introduction}}
\label{sec:introduction}



Painting classification is used in the art history domain for knowledge discovery through object and pose detection in paintings. It also has other uses in style and technique identification through statistical analysis or image similarity along with artist identification.
It is challenging to train classifiers on painting collections due to model bias from domain gaps and data bias from the uneven distribution of artistic styles. Previous techniques like data distillation, traditional and data augmentation improve classifier training using task-specific training datasets or domain adaption. We propose a system to handle data bias in small paintings datasets like the Kaokore dataset \cite{tian2020kaokore} while accounting for domain adaptation in finetuning a model trained on real-world images. Our system comprises two stages: style transfer, and classification. During style transfer, we generate the stylized training samples per class while training the style transformation network's decoder to the training dataset's domain. At classification, we can interpret the effectiveness of the style and content layers at the attention layers when training on the original training dataset and the stylized images. We achieve comparable results to the state-of-the-art (SOTA) with fewer training epochs and classifier parameters.

\begin{figure}[!h]
  \centering
   \includegraphics[width=0.8\linewidth]{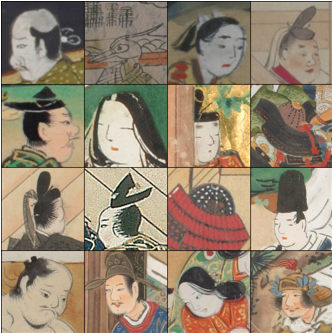}
  \caption{Image samples from the Kaokore dataset.}
  \label{fig:kimgs}
 \end{figure}


Previous work has tried to solve data efficiency in model training for small and uneven datasets in a variety of ways. Data distillation and condensation techniques have opted to create a synthetic dataset that is optimal for the model \cite{zhao2021DC,li2020dada,zhao2020dataset,wang2018dataset}. Although it provides a compressed representation of the training dataset, it overfits to a task distribution. Traditional data augmentation techniques use heuristics to select transformations on their training data \cite{48557,49393} such that the synthetic data belong to the training distribution. However, these do not account for domain adaptation when fine-tuning models, reducing sampling bias solely for the training data. As a possible solution, the model's learned features account for the source data using techniques such as style transfer. It adapts the style from one input image while preserving the content or structure in the second image, using the style and content information from the model's features. Style transfer data augmentation techniques \cite{10.1007/978-3-030-84529-2_7,9577786,jackson2019style,zheng2019stada,9716108} transfer the style information from the target to the source for domain generalization through style invariance. The classification performance can vary from the choice of the style image, with the style set determining the class of augmentations. The model can learn faster with augmentations tailored to the learning algorithm. 



Although data augmentation techniques have been used to improve classifier training, domain adaptation or solve data bias in class imbalance, they treat them as independent problems to solve at either the data level or the model level. Our work aims to utilize the strength of style transfer to tailor the data to the domain as learned from the backbone of the model to create a data augmentation that changes the data's style and content in the perspective of the model's features to help training as well as consider domain adaptation. By producing style transfer augmentations of different proportions for the majority and minority classes, we can select the styles for different parts of the data distribution for classes with different amounts of data. The augmentations for the minority class form the rare samples, while those of the majority class form the representative samples.


In this paper, we propose a system that solves the problems through the stages of transforming content images into class-preserving stylized images using Style Transfer with AdaIN \cite{huang2017arbitrary} and classifying the model with the original and stylized images. The first stage mitigates data bias by selecting style images that represents the mean or outlier of the cluster, thereby letting the model overfit on the class in the former case and regularizing the model in the latter case. The second stage tailors the stylized images to the data per class with domain specific style transformer decoders. The third stage classifies the model with the augmented and original training data and provides the spatial attention to help identify the data bias at the clustering stage by producing interpretable attention maps.

We conduct a series of experiments to check if class imbalance is mitigated through qualitative and quantitative studies. The qualitative studies are the classifier's high and low confidence samples along with the attention map responses for class balancing and the importance of the style and content layers. Through the quantitative studies, we can check the importance of the spatial attention layer and the data augmentation strategy. We achieve comparable results on the Kaokore dataset with the SOTA accuracy score of 89.04\% after 90 epochs using the LOOK method \cite{feng2021rethinking} as compared to our system with 83.22\% after 20 epochs and with a model that requires less training parameters. By changing the proportion of $p_1$ and $p_2$, we can achieve 78.67\% precision and 75.3\% recall with a ResNet-50 \cite{8745417} backbone. We analyze trends from different proportions of augmentations for the majority and minority classes and check its effectiveness for classifiers with different representation capacities. 


Our main contributions include:

\begin{itemize}
    \item We present a spatial attention classification system that achieves comparable results to the SOTA performance from the LOOK model in Kaokore dataset with significantly less training time and training parameters.

    \item We propose to tackle data bias with data balancing using a style transfer based data augmentation method, in which styles are extracted from different levels of deep features.
    
    \item We showcase that we can trade-off accuracy gain versus precision/recall gain by dynamically adjusting the ratio of augmentation between rare and representative classes.
    
    \item Our code is open sourced for validation and further research: \url{https://github.com/41enthusiast/ST-SACLF}
\end{itemize}








\section{\uppercase{Related Work}}

Our work concentrates on painting classification, which is a domain with limited data. Due to this constraint, data efficiency or artificially increasing the amount of training samples can prove beneficial. The training data can improve the model performance by transforming its representation towards the model objective. The section discusses the training data modification at the distribution level, by synthesizing samples at the data or feature level, and at the data level without a model.

\subsection{Data Distribution Manipulation}
Previous works have synthesized data augmentations, modifying the training dataset from the model gradients \cite{zhao2021DC,li2020dada,zhao2020dataset} to condense and distill data into salient model representations. Data distillation techniques \cite{wang2018dataset} have the advantage of providing a reduced yet efficient representation of the training data.  

These techniques summarize the training distribution into a representation that is tailored towards the model or a shared embedding space between the training and target data distribution. The proposed work learns a class-wise transformation for each image from model layer embeddings. It focuses on mitigating data bias through style invariance rather than compression.

\subsection{Style Transfer for Data Augmentation}
Style transfer for data augmentation can aid classification at the data or feature level. Previously, style transfer techniques were slow, iterative optimization techniques \cite{gatys2015neural} that modified the stylized image while leaving the model layers untouched. The transferred style also does not align with the content. However, since the model has a relaxed objective of style invariance, content-specific style transfer is not a priority. Later techniques \cite{huang2017arbitrary,Chandran_2021_CVPR,kolkin2022neural} included a separate transformation network that could be used in inference to generate the stylized images. 

At the data level, style transfer modifies the training distribution itself, whereas at the feature level, it modifies the model's features. Smart Augmentation uses the model features to blend samples selected from strategies like clustering \cite{7906545} to generalize learned augmentation strategies from one network to another. Style transfer similarly blends images corresponding to model features for the style and content. STDA-inf \cite{10.1007/978-3-030-84529-2_7} augments the training data pool with the variations interpolated between intraclass or interclass specific styles and the average of all styles during training. StyleMix and StyleCutMix \cite{9577786} explores the degree of the effect of style and content in the synthetic samples and assign the mixed data a label based on the ratio of the source images. Style Augmentation \cite{jackson2019style} and  STADA \cite{zheng2019stada}, explore the technique effectiveness with different degrees of style in the stylized image for model robustness. STDA-inf and StyleMix are very closely tied to our work, but they do not address the problem of class imbalance.  

 At the feature level, style transfer at the model's feature maps helps in domain generalization as well as  model robustness \cite{9716108}. It generates feature maps across multiple source domains for the feature extractor by injecting style as noise in the layers. The original features and augmented features are both used to train the classifier.

\subsection{Model Agnostic Data Augmentation}
Model agnostic data augmentation techniques modify the training data independently or interdependently \cite{48557,49393} involving only the training data itself. MixUp is an image blending technique with the samples either selected at random or according to the model. The training data in MixMatch is independently processed by applying traditional image augmentation techniques like rotations, normalization, adding or removing noise, recolorization along with geometric operations like shearing and translation.

The choice for the augmentation can also be learned to utilize the model's inductive bias \cite{47890,wang2017effectiveness}. A style transformation network using GAN \cite{wang2017effectiveness} achieves this using meta learning by learning augmentations on a small network that generalize to a larger network. Autoaugment \cite{47890}, on the other hand, uses policies from reinforcement learning to select augmentations. The policy based augmentations are retrieved from sampling a selection pool consisting of traditional image augmentations. The selected augmentations are indicative of domain level knowledge and induce bias based on the model architecture. Our system operates at the data level by randomly samples styles from the same class to preserve the intraclass distribution and mitigate sampling bias by adding more data to each class in different amounts.

Our system also differs from our competitors that use contrastive learning \cite{islam2021broad,feng2021rethinking}, that utilizes the similarity and differences in data to improve model training efficiency, to train all of their model parameters. Contrasting our competitors, our classifier backbone consist of pretrained models \cite{canziani2016analysis,8745417} that were trained on another task with its head finetuned for paintings classification.


\section{\uppercase{Methodology}}
The current data augmentation techniques do not consider how to mitigate class imbalance in interclass settings while giving the option to focus on improving performance or mitigating bias. Neither does the style transfer based data augmentations tune the style and content to the task. 

Our proposed system seeks to address the above issues by the following system features:
\begin{itemize}

    \item We can reduce data bias or promote model performance by adding different proportions of style transfer augmented data to the majority and minority classes. Style transfer augmentations also promote texture invariance through multiple styles per sample, forcing the model to focus on the image content.
    
    \item We make the level of details from style transfer layers configuration to be inline with the model classification through spatial attention modules. These increase the contribution of local level features to the classification loss, thereby reducing the difference in model performance from data augmentations with different style transfer configurations.

\end{itemize}

\begin{figure}[!ht]
  \centering
   \includegraphics[width=0.6\linewidth,height=0.4\textheight]{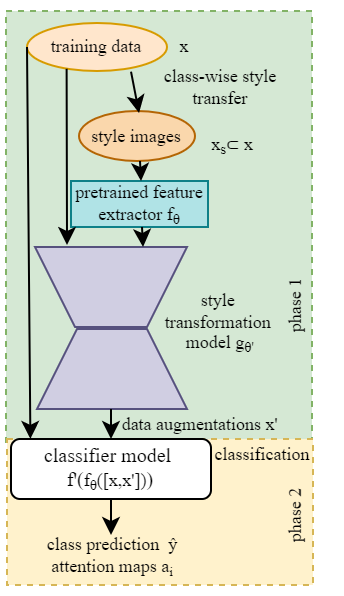}
  \caption{The overall system for style based data augmentation to improve model classification. }
  \label{fig:overall}
 \end{figure}

The system consists of two parts as shown in Figure \ref{fig:overall}. The style transfer transforms the training data into their data augmented counterparts. For each transformation, it uniformly samples a random pair of content and style images from a class to form hybridized samples. Finally, the original and augmented datasets feed into a classifier with a pre-trained network and a head trained on the combination of local and global spatial attention modules. The style transfer uses the same VGG-19 backbone, while the classifier can have different pre-trained backbones.

\subsection{Data Augmentation from Style Transfer}
An automatic method of selecting style images compared to STaDA \cite{zheng2019stada} can remove the subjectivity in selecting style images.
\begin{figure*}[!ht]
  \centering
   \includegraphics[width=\textwidth]{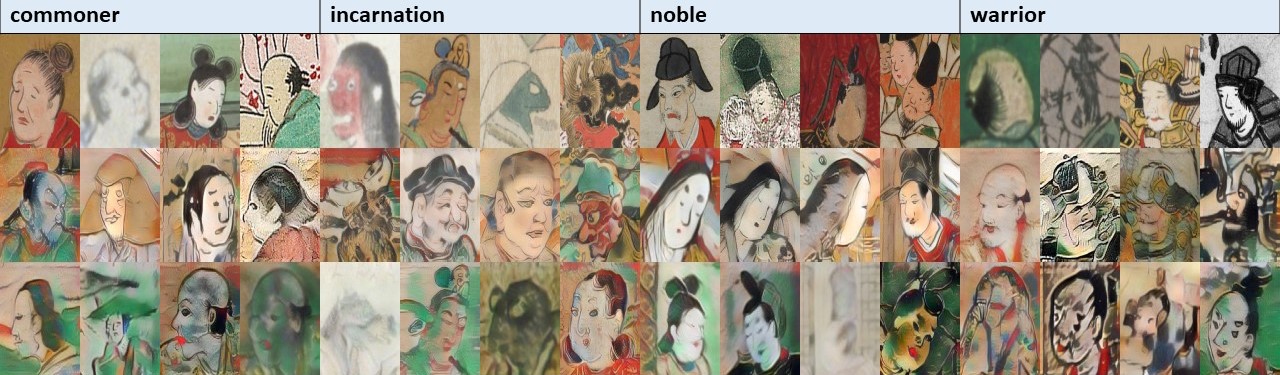}
  \caption{The original samples per class followed by good and sub-optimal style transfer augmentations in the second and third rows, respectively.}
  \label{fig:augs}
 \end{figure*}

We propose to use Adaptive Instance Normalization's \cite{huang2017arbitrary} image transformation network for fast transformation speed with certain flaws. The stylized image would not align the transferred textures from the style image to the content image since it is not context aware. The transformation network is also configuration specific in the resultant textures and is dependent on a specially trained VGG-19 backbone.

\begin{figure*}[!ht]
  \centering
   \includegraphics[width=.8\linewidth]{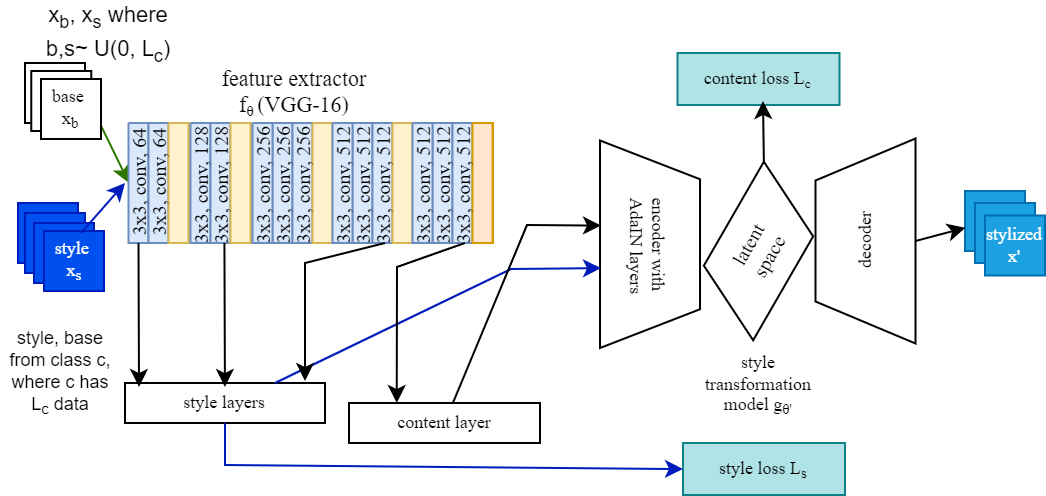}
  \caption{The style transfer model generates stylized versions of the input data per class.}
  \label{fig:st}
 \end{figure*}

Style transfer could account for the difference in domains from the original training dataset with real-world images compared to paintings.  These differences can range from low-level details like texture and pattern differences along with stroke level information to that in the high level like different shapes. By providing style invariance, we can reduce this large domain gap that can create problems in fine-tuning and data generalization \cite{yosinski2014transferable}. We can utilize style transfer to obfuscate the dataset's style and distortions, thereby reducing the domain gap during transfer learning. The classifier is forced to utilize the content information that is common to both the source and target datasets considering that convolutional neural networks are more sensitive to texture information \cite{von2021self}. Data augmentations can separate the content information that would be shared across these real and abstracted depictions, allowing for the higher-level features to be better utilized for classification \cite{geirhos2018imagenettrained}. \cite{10.1109/TMM.2020.3009484} corroborates with the usefulness of learning style invariance while bypassing artistic semantics like brush details, and pattern densities.

We present the data augmented counterparts to the training data that are generated pre-training and using one model itself unlike Smart Augmentation \cite{7906545}. The data augmentation method neither requires an encoder like GANs to exaggerate the details at the chosen feature levels nor does it need a separate network to train augmentation strategies for the main classification network.

The style transfer model, from Figure \ref{fig:st}, optimizes the style loss with the gram matrix of its feature embeddings to account for second-order statistics corresponding to texture and feature variance. The content loss is computed at the bottleneck of the image transformation model to incorporate the style modulation at the Adaptive Instance Normalization \cite{huang2017arbitrary} layers with the content from the reconstruction loss to train the decoder end of the transformation model. It uses a modified pre-trained VGG-19 model with normalized weights as the encoder. We train the style transfer model on uniformly sampled style data from the entire dataset to expose the model to more style varieties. Once the decoder has been trained on the training images in a domain, the style transfer can be computed quickly at inference with uniformly sampled content and style images with repetition per class. AdaIN is a technique that modulates the mean and covariance of the content feature map to that of the style feature map, thereby fusing the information from both inputs.

\begin{equation}
\begin{aligned}\label{eq1}
    &c = f(x_b)\\
    &s = f(x_s)\\
    &AdaIN(c,s)=\sigma(s)\left(\frac{c-\mu(c)}{\sigma(c)}\right)+\mu(s)\\
    &t = AdaIN(c,s)
\end{aligned}
\end{equation}

where c and s are content and style features from the feature extractor, respectively. $\sigma$ is the variance and $\mu$ is the mean, respectively. $t$ is the AdaIN output. It modulates the content feature by the style statistics  at the style transformation network's encoder.

The content loss $L_c$ and the style loss $L_s$ are given as MSE losses and are computed as follows:

\begin{equation}
    \begin{aligned}
    \label{contentandstyleloss}
    &L_c = ||f(g(t) - t)||_2\\
    &L_s = || \mu(\phi_i(g(t))) - \mu(\phi_i(x_s))||_2 +\\
            &\sum_{i=1}^L || \sigma(\phi_i(g(t))) - \sigma(\phi_i(x_s))||_2
    \end{aligned}
\end{equation}

where $t$ is the AdaIN output from Equation \ref{eq1} and content target, $x_s$ is the style image, $f$ is the encoder, $g$ is the decoder, $\phi_i$ are the style layers. The style loss matches the mean and standard statistics between the style image and the stylized image. The content loss matches the stylized features to the target features.

During style transfer, only the weights of the decoder are updated in the training process. After encoding the style and content features for their respective selected layers, they are used to create a stylized tensor using the AdaIN layer. The stylized tensor can retain more information from the style or the structure information depending on the alpha value. It is passed through the decoder to form a hybrid image that retains its structure information by matching its content embedding against the stylized tensor using the content loss. It retains the style information by matching its style embeddings against that of the style image using the style loss. These two losses influence the hybrid image learned by the decoder.

Figure \ref{fig:augs} shows the quality of the generated samples per class. Since most of the images are face-centered, the resultant style transfer transfers the texture while preserving the content. However, since there is no constraints on the contents transferred, some colors bleed into the stylized images as shown in the bottom row. In the Kaokore dataset, there are a lot of green backgrounds and characters with green clothing, it is the common color that bleeds into the samples.

\subsection{Spatial Attention based Image Classifier}

\begin{figure}[!ht]
  \centering
   \includegraphics[width=0.42\textwidth]{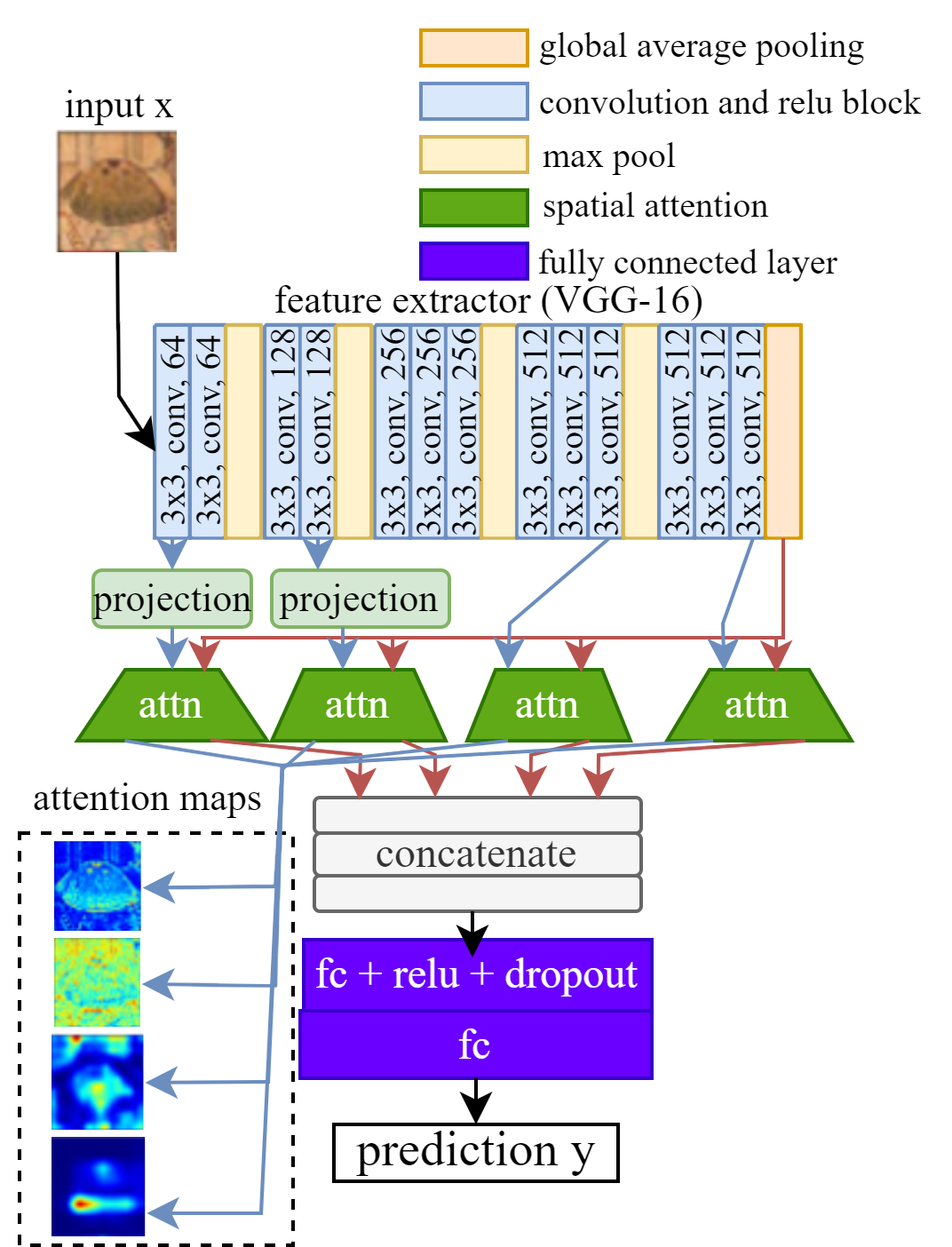}
  \caption{The classifier architecture is depicted with the model flow from the input to the outputs. The blue line indicates local features while the red line indicates global features. The output from the spatial attention layer to the fully connected layers are global features weighted by the corresponding local features.}
  \label{fig:classifier}
 \end{figure}

The classifier, depicted in Figure \ref{fig:classifier}, is made from a pre-trained image classification model like VGG-16 and ResNet-50 \cite{canziani2016analysis,8745417} followed by extracting the very first layer and selecting 3 layers between the first and last layers to correspond to features with more spatial information to represent richer features and create a balance between the style and content information's contribution to the classification loss. The spatial attention module takes the re-projected layer for computing attention with the global feature from the bottle neck. They are concatenated and passed to the head with dense layers and dropout for image classification. It has no batch norm layer and has no global training statistics that can be re-utilized at test, with previous work utilizing only these statistics to account for domain adaptation \cite{frankle2020training}. In this manner, the data augmentation can account for the domain adaptation in the model. With the proposed work, we explore a model agnostic way of domain adaptation and mitigating data bias resulting from the Kaokore dataset's class imbalance. 

Spatial attention can both help in visualizing the impact of style transfer as well as remember coarse to fine detail present in the image. The learnt attention map is further biased since the input data is already amplified by the selected layers. It serves as both a weak supervision signal \cite{jetley2018learn} and the attention mechanism acts as a pseudo memory bank for context retention among the features fed to the module. 

The spatial attention module computes the attention map for the local response map and the global feature at the end of the feature extractor. This embeds both the local and global context of the image. When processing the concatenated spatial attention responses at the MLP head, the style transfer layers are prioritized in the loss computation.

Focal loss is the classification loss used for the spatial attention classifier to help mitigate class imbalance and is formulated as:

\begin{equation}\label{focal_loss}
    \begin{aligned}
    &p_t = softmax(y_{pred})\\
    &softmax(y_{pred}) = \frac{\exp^{y_{pred}}}{\sum_{j=1}^{c}\exp^{{y_{pred}}_j}}\\
    &FL(p_t) = -\alpha(1-p_t)^\gamma y\log(p_t)\\
    \end{aligned}
\end{equation}

In the eqn \ref{focal_loss}, $\alpha$ and $\gamma$ are hyperparameters that can be tuned according to the level of class imbalance in the problem, with higher values for more skewed datasets with more false positives. We can get $p_t$ by passing a softmax function to the logits output $y_{pred}$ from our spatial attention classifier with $c$ as the number of classes. $y$ is the target one hot vector and $p_t$ is the predicted probability.

\begin{figure}[!ht]
  \centering
   \includegraphics[width=0.8\linewidth]{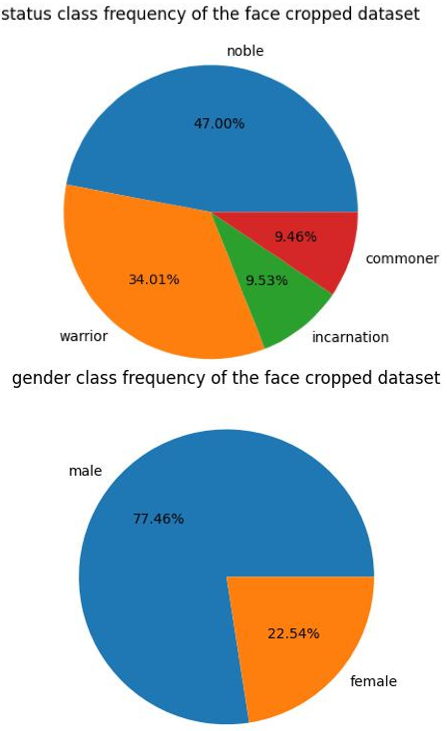}
  \caption{Class imbalance in the Kaokore dataset.}
  \label{fig:kaokore-ds}
 \end{figure}

\section{\uppercase{Experiments}}

We depict different experiments with our system as follows. Section \ref{subsec:datasets} describes the Kaokore dataset which are used in the experimentations at Sections \ref{subsec:qualres} and \ref{subsec:ablationstudies}. The qualitative experiments (Section \ref{subsec:qualres}) explore the interpretability of the style and content layers, while the quantitative experiments are done with ablation studies (Section \ref{subsec:ablationstudies}) to check the effectiveness of the system modules, the classifier and type of data augmentation. Finally, Section \ref{subsec:implementationdets} describes the system configuration.
 
\subsection{Datasets}\label{subsec:datasets}
 
The Kaokore dataset \cite{tian2020kaokore} is a collection of Japanese paintings categorized in two ways according to gender and status. It provides diverse faces within and between classes with different shapes, poses and colors. Thus, it makes a suitable choice for improving classification under style invariance. The gender categorization has the male and female subclasses, while status is subdivided into commoner, noble, incarnation or non human or avatar and warrior. It is very class imbalanced as indicated in Figure \ref{fig:kaokore-ds} and it consists of face cropped images as seen in Figure \ref{fig:kimgs}. The results will be mainly focused on the status to better showcase the impact of style transfer in classification since it requires more finesse than hyperparameter tuning and model regularization techniques unlike the gender classification task. The dataset is fairly small with 6,756 training images, 845 validation and test images of the same size and could benefit from transfer learning.



\begin{figure}[!ht]
 \begin{subfigure}[b]{0.5\textwidth}
\begin{subfigure}[b]{\textwidth}
   \includegraphics[width=\linewidth]{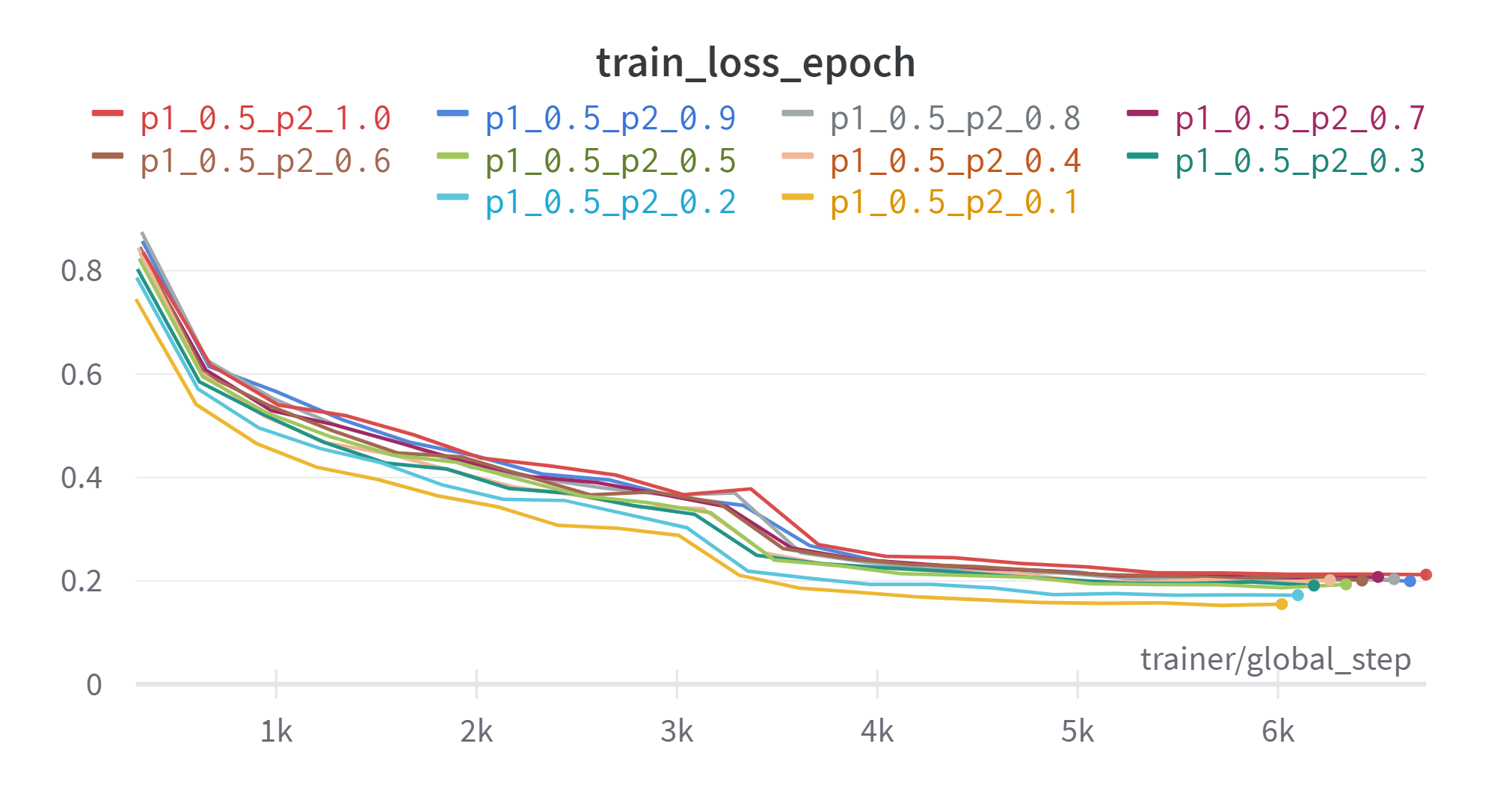}
  \caption{Training convergence with differing amounts of rare samples ($p_2$)}
  \vspace{1mm}
  
 \end{subfigure}
 
   \includegraphics[width=\linewidth]{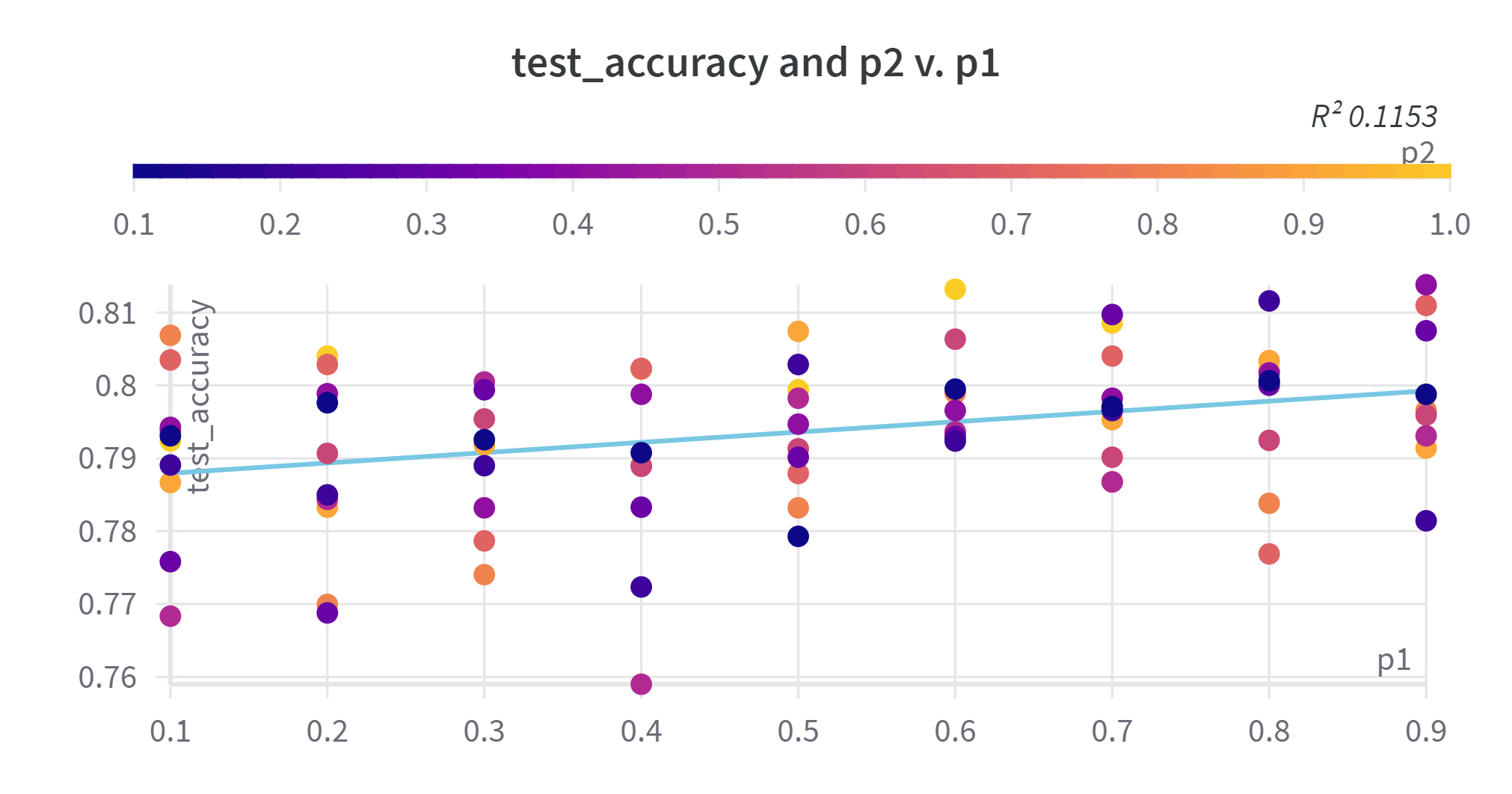}
  \caption{Test accuracy scores with differing amounts of rare ($p_1$) and representative ($p_2$) augmentations}
  \vspace{1mm}
 \end{subfigure}

 \begin{subfigure}[b]{0.5\textwidth}
   \includegraphics[width=\linewidth]{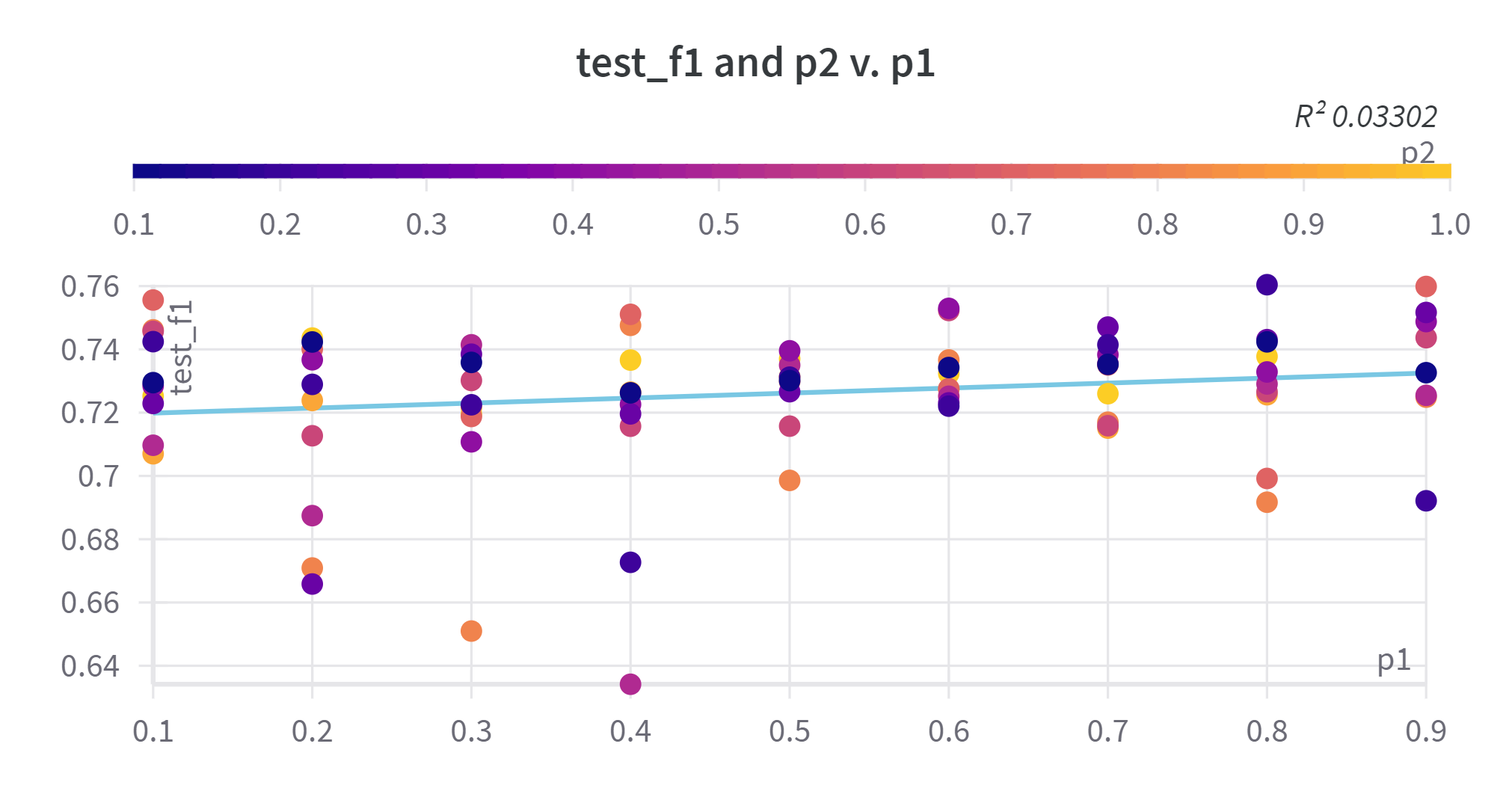}
  \caption{Test F1 scores with differing amounts of rare ($p_1$) and representative ($p_2$) augmentations}
  \vspace{1mm}
 \end{subfigure}
 \caption{Model performance trends with differing amounts of style transfer augmentation. $p_1$ and $p_2$ indicate percentages of data in the common classes, noble and warrior, and rare classes, incarnation and commoner, used as extra training data.}
 \label{fig:trends}
\end{figure}

\subsection{Quantitative Results}
\label{subsec:ablationstudies}

The following experiments were conducted to test the efficacy of style transfer as a data augmentation technique. The first is an analysis of the style transfer effects on models of different capacities and architectures. The second explores the model performance under different configurations of $p_1$ and $p_2$. This is followed by a comparison with state-of-the-art methods.



\begin{table}[ht]
\centering
\resizebox{0.47\textwidth}{!}{
\begin{tabular}{|l|l|llll|}
\hline
\multirow{2}{*}{\begin{tabular}[c]{@{}l@{}}Model \\ Architecture \end{tabular}} &
  \multirow{2}{*}{\begin{tabular}[c]{@{}l@{}}Style transfer data \\ augmentation type\end{tabular}} &
  \multicolumn{4}{l|}{Metrics (in percentage)} \\ \cline{3-6} 
                          &                                  & \multicolumn{1}{l|}{Accuracy} & \multicolumn{1}{l|}{Recall} & \multicolumn{1}{l|}{Precision} & F1 score \\ \hline
\multirow{2}{*}{VGG16}    & \begin{tabular}[c]{@{}l@{}}Optimal rare and \\ representative mix\end{tabular} & \multicolumn{1}{l|}{82.06}    & \multicolumn{1}{l|}{71.41}  & \multicolumn{1}{l|}{75.9}     & 73.27    \\ \cline{2-6} 
                          & No augmentation                          & \multicolumn{1}{l|}{79.91}         & \multicolumn{1}{l|}{71.08}       & \multicolumn{1}{l|}{73.09}          & 72.00         \\ \hline
\multirow{2}{*}{VGG19}    &\begin{tabular}[c]{@{}l@{}}Optimal rare and \\ representative mix\end{tabular} & \multicolumn{1}{l|}{80.68}    & \multicolumn{1}{l|}{71.43}  & \multicolumn{1}{l|}{74.06}     & 72.39    \\ \cline{2-6} 
                          & No augmentation                          & \multicolumn{1}{l|}{78.84}         & \multicolumn{1}{l|}{67.67}       & \multicolumn{1}{l|}{73.34}          & 69.80         \\ \hline
\multirow{2}{*}{ResNet34} & \begin{tabular}[c]{@{}l@{}}Optimal rare and \\ representative mix\end{tabular} & \multicolumn{1}{l|}{81.38}    & \multicolumn{1}{l|}{73.83}   & \multicolumn{1}{l|}{76.40}     & 74.88    \\ \cline{2-6} 
                          & No augmentation                          & \multicolumn{1}{l|}{80.03}         & \multicolumn{1}{l|}{71.49}       & \multicolumn{1}{l|}{75.93}          & 73.29         \\ \hline
\multirow{2}{*}{ResNet50} & \begin{tabular}[c]{@{}l@{}}Optimal rare and \\ representative mix\end{tabular} & \multicolumn{1}{l|}{\textbf{83.22}}    & \multicolumn{1}{l|}{\textbf{73.86}}   & \multicolumn{1}{l|}{\textbf{76.9}}     & \textbf{75.2}    \\ \cline{2-6} 
                          & No augmentation                          & \multicolumn{1}{l|}{78.43}         & \multicolumn{1}{l|}{69.55}       & \multicolumn{1}{l|}{71.58}          & 70.48         \\ \hline
\end{tabular}}
\caption{Model performance for different classifier backbones with and without data augmentation.}
\label{tab:example1}
\end{table}

\begin{table*}[ht]
\centering
\resizebox{\textwidth}{!}{
\begin{tabular}{|l|l|l|l|l|l|l|l|l|l|l|}
\hline
p1/p2 &
  0.1 &
  0.2 &
  0.3 &
  0.4 &
  0.5 &
  0.6 &
  0.7 &
  0.8 &
  0.9 &
  1.0 \\ \hline
0.1 &
  79.31/76.89/70.55 &
  78.91/75.46/73.23 &
  77.58/76.33/69.54 &
  79.42/74.9/71.5 &
  76.83/73.9/69.43 &
  79.37/76.03/73.34 &
  80.35/75.96/75.21 &
  80.69/76.44/73.23 &
  78.67/73.1/69.13 &
  79.24/73.61/71.61 \\ \hline
0.2 &
  79.76/75.93/72.82 &
  78.5/75.36/71.26 &
  76.88/72.01/64.97 &
  79.89/74.56/72.92 &
  78.44/74.16/65.81 &
  79.07/72.84/70.49 &
  80.29/75.77/72.66 &
  76.99/70.92/65.76 &
  78.33/74.4/70.89 &
  80.4/75.47/73.41 \\ \hline
0.3 &
  79.26/74.29/72.97 &
  78.9/73.75/71.04 &
  79.94/74.54/73.24 &
  78.32/73.97/69.23 &
  80.05/74.75/73.61 &
  79.54/74.75/73.61 &
  79.54/74.64/71.91 &
  77.87/73.35/70.67 &
  77.4/70.26/64.99 &
  79.19/73.93/71.1 \\ \hline
0.4 &
  79.08/75.22/70.69 &
  77.23/74.38/65.6 &
  78.33/73.23/70.98 &
  79.88/75.09/70.38 &
  75.9/72.43/61.49 &
  78.89/72.41/71.15 &
  80.24/76.94/73.6 &
  80.23/76.05/73.82 &
  78.91/73.98/71.69 &
  80.23/75.69/72.14 \\ \hline
0.5 &
  77.93/74.17/72.21 &
  80.29/76.2/71.16 &
  79.02/73.37/72.14 &
  79.47/76.59/71.95 &
  79.83/75.52/72.04 &
  79.13/73.64/70.11 &
  78.79/73.61/72.4 &
  78.32/72.86/68.59 &
  80.74/76.19/71.37 &
  79.94/74.82/72.98 \\ \hline
0.6 &
  79.95/76.04/71.71 &
  79.24/73.31/71.28 &
  79.3/74.03/70.97 &
  79.66/76.91/73.98 &
  79.36/73.41/71.73 &
  80.64/77.58/73.53 &
  79.36/76.22/70.69 &
  79.9/75.16/72.69 &
  79.25/73.7/71.91 &
  81.32/76.04/71.52 \\ \hline
0.7 &
  79.71/74.61/72.64 &
  79.66/76.4/72.43 &
  80.97/76.59/73.26 &
  79.83/74.28/73.45 &
  78.68/75.03/72.38 &
  79.02/75.24/69.63 &
  80.4/75.64/72.17 &
  78.68/73.88/70.29 &
  79.53/73.71/70.04 &
  80.86/75.89/71.19 \\ \hline
0.8 &
  80.07/76.38/72.7 &
  81.16/77.31/\textbf{75.03} &
  80/75.29/73.46 &
  80.17/74.82/72.12 &
  80.05/76.12/70.87 &
  79.25/75.32/70.7 &
  77.69/71.7/68.91 &
  78.38/73.22/66.91 &
  80.34/74.29/71.6 &
  80.23/75.3/72.74 \\ \hline
0.9 &
  79.88/75.88/71.76 &
  78.14/72.34/69.05 &
  80.75/77.13/73.84 &
  \textbf{81.38}/76.4/73.83 &
  79.31/74.65/71.06 &
  79.59/75.63/73.29 &
  81.1/77.7/74.61 &
  79.66/73.98/71.43 &
  79.14/74.34/70.97 &
  79.08/74.85/71.52 \\ \hline
1.0 &
  77.87/73.32/66.63 &
  79.78/76.16/73.5 &
  78.15/73.47/70.27 &
  79.71/74.37/70.57 &
  79.36/74.72/70.42 &
  78.15/77.43/62.81 &
  80.41/\textbf{78.62}/73.6 &
  80.24/76.39/71.44 &
  80.29/76.37/73.45 &
  78.96/74.33/69.42 \\ \hline
\end{tabular}}
\caption{ResNet-34 model performance metrics (accuracy/precision/recall) for different p1 and p2 configurations, where p1 is the percentage of extra majority class data and p2 is the percentage of extra minority class data.}
\label{prob_sweep}
\end{table*}

\begin{table}[ht]
\centering
\scriptsize
\begin{tabular}{|l|l|l|l}

\cline{1-3}
Method                                                                         & \begin{tabular}[c]{@{}l@{}}Test \\ accuracy\end{tabular} & \begin{tabular}[c]{@{}l@{}}Number of \\ trainable \\ parameters \\ (in millions)\end{tabular}                     &  \\ \cline{1-3}
\begin{tabular}[c]{@{}l@{}}VGG-11 \cite{tian2020kaokore}\end{tabular}    & 78.74\%                                                  & 9.2 M                                                                                          &  \\ \cline{1-3}
\begin{tabular}[c]{@{}l@{}}AlexNet \cite{tian2020kaokore}\end{tabular}    & 78.93\%                                                  & 62.3 M                         

&  \\ \cline{1-3}
\begin{tabular}[c]{@{}l@{}}DenseNet-121 \cite{tian2020kaokore}\end{tabular}    & 79.70\%                                                  & 7.6 M                                                    
&  \\ \cline{1-3}
\begin{tabular}[c]{@{}l@{}}Inception-v3  \cite{tian2020kaokore}\end{tabular}    & 84.25\%                                                  & 24 M                                                    

&  \\ \cline{1-3}
\begin{tabular}[c]{@{}l@{}}ResNet-18 \cite{tian2020kaokore}\end{tabular}    & 82.16\%                                                  & 11 M                                                                                          &  \\ \cline{1-3}
\begin{tabular}[c]{@{}l@{}}MobileNet-v2  \cite{tian2020kaokore}\end{tabular}    & 82.35\%                                                  & 3.2 M                                                                                          &  \\ \cline{1-3}
\begin{tabular}[c]{@{}l@{}}ResNet-34 \cite{tian2020kaokore}\end{tabular} & 84.82 \%                                                 & 21.3 M                                                                                                                &  \\ \cline{1-3}
SelfSupCon \cite{islam2021broad}                                                                       & 88.92\%                                                  & 47 M                                                                                          &  \\ \cline{1-3}
CE+SelfSupCon \cite{islam2021broad}                                                                     & 88.25\%                                                  & 27.9 M                                                                                        &  \\ \cline{1-3}

LOOK (ResNet-50) \cite{feng2021rethinking}                                                                          & \textbf{89.04} \%                                                   & 23.5 M                                                                                                                     &  \\ \cline{1-3}

\textbf{Ours (VGG-16 backbone) }                                                                    & 82.06 \%                                                 & \textbf{1.2 M }                                                                                                                &  \\ \cline{1-3}
\textbf{Ours (ResNet-34 backbone)  }                                                           & 81.38 \%                                                 & \textbf{1.2 M}                                                                                                                 &  \\ \cline{1-3}
\textbf{Ours (ResNet-50 backbone)  }                                                           & 83.22 \%                                                 & 20.1 M                                                                                                                 &  \\ \cline{1-3}

\end{tabular}
\caption{A comparative study for the Kaokore dataset. Note that our data augmentation method can be used on top of all existing state-of-the-arts and boost their performance.}\label{table3}
\end{table}

Style transfer has better results when the model capacity is larger, as seen in VGG-19 and ResNet-34 in Table \ref{tab:example1}. The control case in the tables are the models that are trained with no data augmentation. The data augmentation type in the table refer to the case when the model is fed the listed types of style transfer transformed training data. It can be inferred that larger models that overfit to the dataset can benefit from style transfer as a type of model regularization. The rare augmentations work better for models with larger capacities since it offers more visual variation while making it harder for the model to overfit on the dataset. In models with smaller backbones like VGG-16 and VGG-19, the representative samples offer better augmentations, since the excessive visual variations in styles can hurt the model performance as seen in \cite{zheng2019stada}. 

Changing the proportions of the data augmentations for the rare and representative samples show trends in model convergence and performance, as seen in Figure \ref{fig:trends}. $p_1$ and $p_2$ are a percentage of the data in majority and minority classes that are used as extra training data, allowing for stratified sampling. The test was performed on the spatial attention classifier with a ResNet-34 backbone since the capacity of larger models benefit from more training data. The model's training convergence is faster with less rare samples and there is a consistent trend for different fixed p1 values. The test accuracy increases with more representative samples and rare samples. On the other hand, the F1 scores mostly benefit from having lesser proportions of rare augmentations than representative augmentations. This trend allows for a trade-off between F1 score, to represent both precision and recall, and accuracy. It also allows for a trade-off between model convergence and potentially overfitting to that of regularization from the added rare samples.

ResNet-50 gets the best performance improvement from the control with no augmentation with $p_1 = 0.3$ and $p_2 = 0.2$, which can be attributed to its increased capacity of 20 million trainable parameters (mentioned in Table \ref{table3}). It has better accuracy with less rare and representative sample proportions as compared to the previous configurations. The larger number of learnable parameters lets the model benefit from more rare samples since it is prone to overfitting on smaller datasets. 

Table \ref{prob_sweep} details the model performance metrics with each cell listing the accuracy, precision and recall respectively. From the Figure \ref{fig:trends} and Table \ref{prob_sweep}, we can infer the choice of the rare proportions ($p_2$) depending on the percentage of extra representative samples ($p_1$).

From the Table \ref{table3}, our best models with the ResNet and VGG Architecture reach comparable results with LOOK \cite{feng2021rethinking} and the five contrastive methods \cite{islam2021broad} with less computation. Our work's competitors all fully finetune their models but we only finetune the head of the classifier. The methods with contrastive learning \cite{islam2021broad,feng2021rethinking} achieve better test accuracy, but they have to be trained longer and have to be completely finetuned to the task. In the settings where they do not do so, they have worse results to our method when they train on a part of the dataset. They also have significantly worse results in a few shot setting, making them both data intensive and computationally expensive. On the other hand, our method is compatible with the SOTA since it is a pretraining step, possibly achieving better results in tandem with their method.






From the tables, on comparing the results from the control to the data augmented case, the performance is more evenly spread out in the latter case, indicating better performance per class from the precision, recall and F1 score metrics. The data augmentations also seem to provide better results than the control for models with more parameters and comparative results for smaller models.

\begin{figure*}[!ht]
 \begin{subfigure}[b]{\textwidth}
   \includegraphics[width=\linewidth]{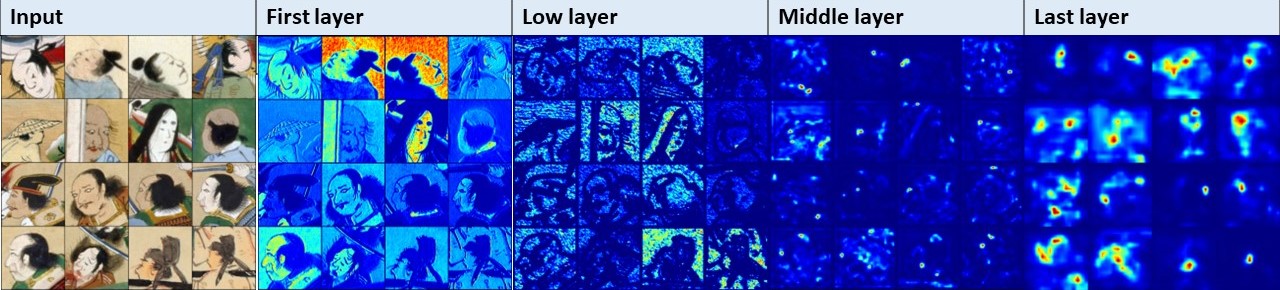}
  \caption{The attention map response without data augmentation for a random test batch.}
  \vspace{1mm}
  \label{fig:amap_noaug}
 \end{subfigure}

 \begin{subfigure}[b]{\textwidth}
   \includegraphics[width=\linewidth]{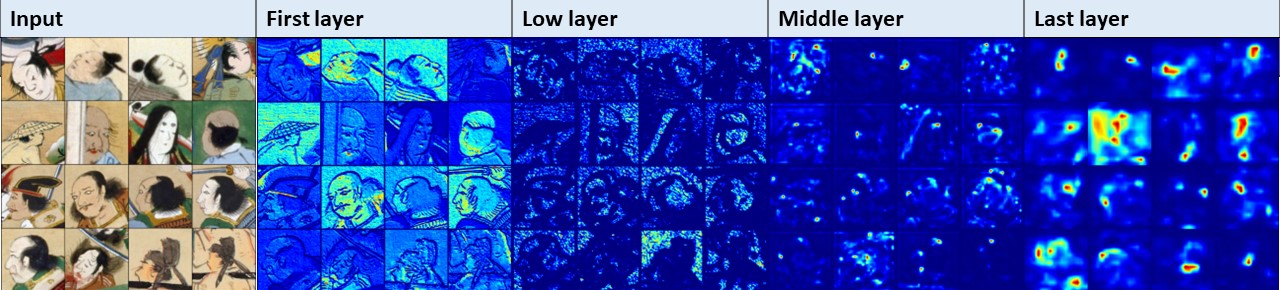}
  \caption{The attention map response with data augmentation for a random test batch.}
  \vspace{1mm}
  \label{fig:amap_aug}
 \end{subfigure}

  \caption{Attention map responses for the style transfer layers in a ResNet architecture. From left to right, they represent the input images, the lowest, low, middle and end layers. The response levels go from low to high and are indicated from dark blue to red.}
  \label{fig:amap}
 \end{figure*}

   \begin{figure*}[!ht]
  \centering
   \includegraphics[width=\linewidth]{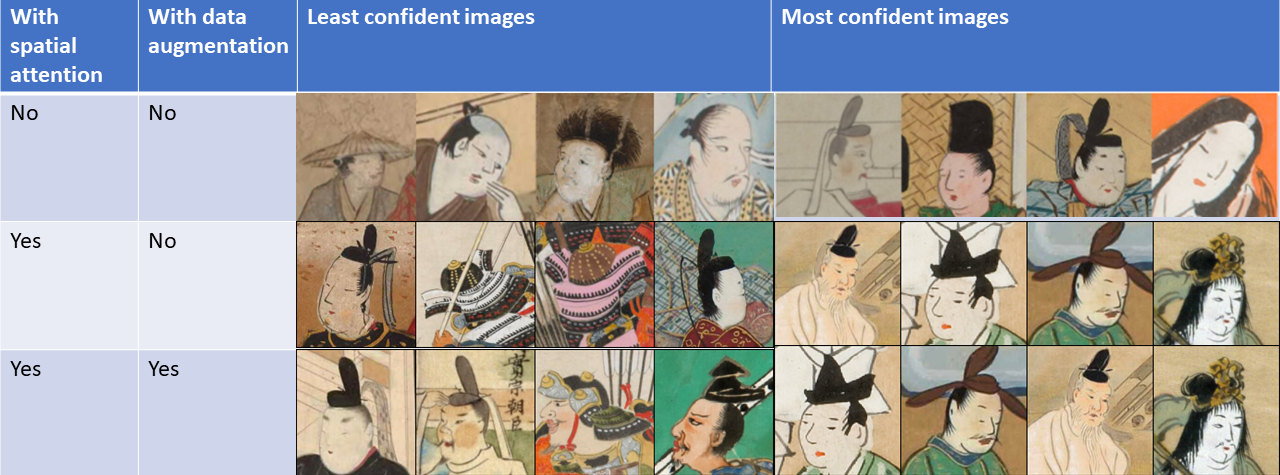}
  \caption{The most and least confident images from the validation subset of the Kaokore dataset for different system configurations in the classifier with a VGG-16 backbone.}
  \label{fig:confident_imgages}
 \end{figure*}

\subsection{Qualitative Results}\label{subsec:qualres}

The visualization of the spatial attention map in Figure \ref{fig:amap} can be used to highlight what parts of the image are considered important to the model's layers. 
As in Figure \ref{fig:amap_noaug}, without data augmentation, the model focuses on a wider area, with higher levels of responses at the lower levels of the model. These lower layers are sensitive to texture, edge and color information. In the Kaokore dataset, the faces can be classified into the different statuses by their hair style and clothes as distinctive features. The faces and certain colors in this case have very high activation responses.

As in Figure \ref{fig:amap_aug}, with data augmentation, we can see the texture details highlighted more than the color information at the lowest level. The regions with faces and background have higher responses and in the later layers, the areas in the vicinity of the hair and subject are given more importance. Overall, there is more levels of activity in the response maps with data augmentation.

The most and least confident images, as seen in Figure \ref{fig:confident_imgages}, provides a check into the classes the model is biased towards. It is formed by ranking the model losses and visualizing the corresponding images. The most confident images have the least losses from left to right, while the least confident images rank the losses in a descending order across the test set. The selection of the Vgg-16 model was motivated by style transfer working better with it as a backbone. The style augmentation version has its least confident images with noble class examples. This could be due to the test set's sampling bias to the noble class. In the first row's configuration, the least confident images are from the commoner class despite its small test sample size, indicating class imbalance. The remaining two configurations have the same images in the most confident images with different rankings. These images have backgrounds with less variation and details. In the case of the system with only spatial attention, the least confident images have complex backgrounds along with subjects with obscured faces. Style transfer based augmentations account for the latter weakness but it does not account for highly complex backgrounds. By providing variations of styles per sample to promote texture invariance, the model could ignore image details when ignoring texture information.

\subsection{Implementation Details}\label{subsec:implementationdets}

During the pre-training phase, the style transfer model is trained on pairs of uniformly sampled style and content images from the training dataset for 20,000 iterations. The learned decoder is used in inference to generate the stylized counterparts by similarly sampling style and content pairs per class. The resultant dataset retains the same distribution as the training dataset, having the same number of samples for each class. It uses the same parameters as the AdaIN style transfer network\cite{huang2017arbitrary}.  

The model is trained with the batch size of 64 and learning rate of 0.0001 for 20 epochs using an Adam optimizer. Additionally, the model uses dropout with a probability of 0.23. It uses L2 regularization along with a focal loss with the gamma and alpha set to 2. Finally, there are 8 workers for faster data processing.

L2 regularization and focal loss facilitates the model to focus parts of the feature, since the style transfer can utilize features of different levels of details that can get lost from the convolution and pooling operations. Dropout was selected to further acerbate this model regularization.

A single NVIDIA A100 GPU instance trained and did inference on the model. It was also used during pretraining to generate data augmented counterparts.

The pre-trained models considered in the classifier are ResNet34, VGG \cite{canziani2016analysis,8745417} and its variants VGG-16 and VGG-19. The ResNet and VGG architectures provide a comparitive study against the benchmarks of the Kaokore dataset \cite{tian2020kaokore}. The VGG variants are used to experiment the effect of the augmentations on the model capacity. Their weights are frozen for all the stages of the system to showcase the strength of data augmentation rather than the model architecture itself. The fully connected layers are removed and the last layer is selected as a global average pooling layer to make the model robust to images of any size and better serve as a feature extractor.

\section{\uppercase{Conclusions}}
\label{sec:conclusion}


We observe that style transfer for data augmentation with the classifier tailored style images and stylization produces better results per class. It also mitigates data bias from class imbalance in small datasets of a different domain. The system achieves this by stylizing images towards the representative and rare clustered samples to bias the classification loss to a changed training manifold. We can balance the tradeoff between accuracy and convergence to recall, precision and f1-score by changing the proportion of extra data per minority and majority class. The amount of extra rare classes to be added range between 20-60\% of the minority classes with more minority classes giving better recall, precision and f1-scores. In the representative classes case, 50-90\% more data can improve all the metrics, with a more pronounced effect on accuracy and model convergence.
We conduct qualitative experiments to check class imbalance and interpretability of the backbone at different layers. Next, we perform quantitative studies to show the weak supervision signal from the spatial attention modules and the reduced data bias through style transfer augmentations. 



While we automate the style images for style transfer through the random sampling of style and content images per class, the learned style space is still subjective due to the variations as a result from the selection of different style and content layers. Future work can look into focused sampling of style and content images to make the style transfer more task oriented. Our work has not experimented with varying the extent of style and content in the image which can also be learned according to suit the task at hand. 

Furthermore, we can use meta learning on top of the system to learn hyperparameters as well as effectively learn the training dataset through the different style transfer augmentations as support sets with fewer samples. Since contrastive learning techniques are highly dependent on the data augmentation techniques, the future work can incorporate it into the model training process. Since the current system allows for flexibility in the choice of model and training pipeline, the style transfer based data augmentation can be adapted in a plug and play manner as a pretraining step. 

Lastly, we will explore the model generalization on other paintings datasets such as PACS \cite{li2017deeper}, WikiArt \cite{saleh2015large} and Rijksmuseum \cite{mensink2014rijksmuseum}. The PACS dataset is a small dataset with subjects portrayed in different media and can be used to check the model’s performance in domain generalization. The WikiArt dataset has paintings of different genres and styles while the Rijksmuseum dataset has a larger collection of data. The two datasets can be used to check the data efficiency of the model with different training data sizes.




\bibliographystyle{apalike}
{\small
\bibliography{main}}

\begin{thebibliography}{}

\bibitem[Berthelot et~al., 2019]{48557}
Berthelot, D., Carlini, N., Goodfellow, I., Papernot, N., Oliver, A., and
  Raffel, C. (2019).
\newblock Mixmatch: A holistic approach to semi-supervised learning.
\newblock In {\em NeurIPS}.

\bibitem[Canziani et~al., 2016]{canziani2016analysis}
Canziani, A., Paszke, A., and Culurciello, E. (2016).
\newblock An analysis of deep neural network models for practical applications.
\newblock {\em arXiv preprint arXiv:1605.07678}.

\bibitem[Carratino et~al., 2020]{49393}
Carratino, L., Cisse, M., Jenatton, R., and Vert, J.-P. (2020).
\newblock On mixup regularization.
\newblock Technical report, arXiv.
\newblock 2006.06049.

\bibitem[Chandran et~al., 2021]{Chandran_2021_CVPR}
Chandran, P., Zoss, G., Gotardo, P., Gross, M., and Bradley, D. (2021).
\newblock Adaptive convolutions for structure-aware style transfer.
\newblock In {\em Proceedings of the IEEE/CVF Conference on Computer Vision and
  Pattern Recognition (CVPR)}, pages 7972--7981.

\bibitem[Cubuk et~al., 2019]{47890}
Cubuk, E.~D., Zoph, B., Mane, D., Vasudevan, V., and Le, Q.~V. (2019).
\newblock Autoaugment: Learning augmentation policies from data.

\bibitem[Feng et~al., 2021]{feng2021rethinking}
Feng, Y., Jiang, J., Tang, M., Jin, R., and Gao, Y. (2021).
\newblock Rethinking supervised pre-training for better downstream
  transferring.
\newblock {\em arXiv preprint arXiv:2110.06014}.

\bibitem[Frankle et~al., 2020]{frankle2020training}
Frankle, J., Schwab, D.~J., and Morcos, A.~S. (2020).
\newblock Training batchnorm and only batchnorm: On the expressive power of
  random features in cnns.
\newblock {\em arXiv preprint arXiv:2003.00152}.

\bibitem[Gatys et~al., 2015]{gatys2015neural}
Gatys, L.~A., Ecker, A.~S., and Bethge, M. (2015).
\newblock A neural algorithm of artistic style.
\newblock {\em arXiv preprint arXiv:1508.06576}.

\bibitem[Geirhos et~al., 2019]{geirhos2018imagenettrained}
Geirhos, R., Rubisch, P., Michaelis, C., Bethge, M., Wichmann, F.~A., and
  Brendel, W. (2019).
\newblock Imagenet-trained {CNN}s are biased towards texture; increasing shape
  bias improves accuracy and robustness.
\newblock In {\em International Conference on Learning Representations}.

\bibitem[Hong et~al., 2021a]{9577786}
Hong, M., Choi, J., and Kim, G. (2021a).
\newblock Stylemix: Separating content and style for enhanced data
  augmentation.
\newblock In {\em 2021 IEEE/CVF Conference on Computer Vision and Pattern
  Recognition (CVPR)}, pages 14857--14865.

\bibitem[Hong et~al., 2021b]{10.1007/978-3-030-84529-2_7}
Hong, T., Zou, Y., and Ma, J. (2021b).
\newblock Stda-inf: Style transfer for data augmentation through in-data
  training and fusion inference.
\newblock In Huang, D.-S., Jo, K.-H., Li, J., Gribova, V., and Hussain, A.,
  editors, {\em Intelligent Computing Theories and Application}, pages 76--90,
  Cham. Springer International Publishing.

\bibitem[Huang and Belongie, 2017]{huang2017arbitrary}
Huang, X. and Belongie, S. (2017).
\newblock Arbitrary style transfer in real-time with adaptive instance
  normalization.
\newblock In {\em Proceedings of the IEEE international conference on computer
  vision}, pages 1501--1510.

\bibitem[Islam et~al., 2021]{islam2021broad}
Islam, A., Chen, C.-F.~R., Panda, R., Karlinsky, L., Radke, R., and Feris, R.
  (2021).
\newblock A broad study on the transferability of visual representations with
  contrastive learning.
\newblock In {\em Proceedings of the IEEE/CVF International Conference on
  Computer Vision}, pages 8845--8855.

\bibitem[Jackson et~al., 2019]{jackson2019style}
Jackson, P.~T., Abarghouei, A.~A., Bonner, S., Breckon, T.~P., and Obara, B.
  (2019).
\newblock Style augmentation: data augmentation via style randomization.
\newblock In {\em CVPR workshops}, volume~6, pages 10--11.

\bibitem[Jetley et~al., 2018]{jetley2018learn}
Jetley, S., Lord, N.~A., Lee, N., and Torr, P.~H. (2018).
\newblock Learn to pay attention.
\newblock {\em arXiv preprint arXiv:1804.02391}.

\bibitem[Kolkin et~al., 2022]{kolkin2022neural}
Kolkin, N., Kucera, M., Paris, S., Sykora, D., Shechtman, E., and
  Shakhnarovich, G. (2022).
\newblock Neural neighbor style transfer.
\newblock {\em arXiv e-prints}, pages arXiv--2203.

\bibitem[Lemley et~al., 2017]{7906545}
Lemley, J., Bazrafkan, S., and Corcoran, P. (2017).
\newblock Smart augmentation learning an optimal data augmentation strategy.
\newblock {\em IEEE Access}, 5:5858--5869.

\bibitem[Li et~al., 2017]{li2017deeper}
Li, D., Yang, Y., Song, Y.-Z., and Hospedales, T.~M. (2017).
\newblock Deeper, broader and artier domain generalization.
\newblock In {\em Proceedings of the IEEE international conference on computer
  vision}, pages 5542--5550.

\bibitem[Li et~al., 2020]{li2020dada}
Li, Y., Hu, G., Wang, Y., Hospedales, T., Robertson, N.~M., and Yang, Y.
  (2020).
\newblock Dada: Differentiable automatic data augmentation.
\newblock {\em arXiv preprint arXiv:2003.03780}.

\bibitem[Mensink and Van~Gemert, 2014]{mensink2014rijksmuseum}
Mensink, T. and Van~Gemert, J. (2014).
\newblock The rijksmuseum challenge: Museum-centered visual recognition.
\newblock In {\em Proceedings of International Conference on Multimedia
  Retrieval}, pages 451--454.

\bibitem[Saleh and Elgammal, 2015]{saleh2015large}
Saleh, B. and Elgammal, A. (2015).
\newblock Large-scale classification of fine-art paintings: Learning the right
  metric on the right feature.
\newblock {\em arXiv preprint arXiv:1505.00855}.

\bibitem[Shah and Harpale, 2018]{8745417}
Shah, U. and Harpale, A. (2018).
\newblock A review of deep learning models for computer vision.
\newblock In {\em 2018 IEEE Punecon}, pages 1--6.

\bibitem[Tian et~al., 2020]{tian2020kaokore}
Tian, Y., Suzuki, C., Clanuwat, T., Bober-Irizar, M., Lamb, A., and Kitamoto,
  A. (2020).
\newblock {KaoKore: A Pre-modern Japanese Art Facial Expression Dataset}.
\newblock In {\em Proceedings of the International Conference on Computational
  Creativity}, pages 415--422.

\bibitem[Virtusio et~al., 2021]{10.1109/TMM.2020.3009484}
Virtusio, J.~J., Tan, D.~S., Cheng, W.-H., Tanveer, M., and Hua, K.-L. (2021).
\newblock Enabling artistic control over pattern density and stroke strength.
\newblock {\em Trans. Multi.}, 23.

\bibitem[Von~K{\"u}gelgen et~al., 2021]{von2021self}
Von~K{\"u}gelgen, J., Sharma, Y., Gresele, L., Brendel, W., Sch{\"o}lkopf, B.,
  Besserve, M., and Locatello, F. (2021).
\newblock Self-supervised learning with data augmentations provably isolates
  content from style.
\newblock {\em Advances in neural information processing systems},
  34:16451--16467.

\bibitem[Wang et~al., 2017]{wang2017effectiveness}
Wang, J., Perez, L., et~al. (2017).
\newblock The effectiveness of data augmentation in image classification using
  deep learning.
\newblock {\em Convolutional Neural Networks Vis. Recognit}, 11:1--8.

\bibitem[Wang et~al., 2018]{wang2018dataset}
Wang, T., Zhu, J.-Y., Torralba, A., and Efros, A.~A. (2018).
\newblock Dataset distillation.
\newblock {\em arXiv preprint arXiv:1811.10959}.

\bibitem[Wang et~al., 2022]{9716108}
Wang, Y., Qi, L., Shi, Y., and Gao, Y. (2022).
\newblock Feature-based style randomization for domain generalization.
\newblock {\em IEEE Transactions on Circuits and Systems for Video Technology},
  32(8):5495--5509.

\bibitem[Yosinski et~al., 2014]{yosinski2014transferable}
Yosinski, J., Clune, J., Bengio, Y., and Lipson, H. (2014).
\newblock How transferable are features in deep neural networks?
\newblock {\em Advances in neural information processing systems}, 27.

\bibitem[Zhao et~al., 2020]{zhao2020dataset}
Zhao, B., Mopuri, K.~R., and Bilen, H. (2020).
\newblock Dataset condensation with gradient matching.
\newblock {\em arXiv preprint arXiv:2006.05929}.

\bibitem[Zhao et~al., 2021]{zhao2021DC}
Zhao, B., Mopuri, K.~R., and Bilen, H. (2021).
\newblock Dataset condensation with gradient matching.
\newblock In {\em International Conference on Learning Representations}.

\bibitem[Zheng et~al., 2019]{zheng2019stada}
Zheng, X., Chalasani, T., Ghosal, K., Lutz, S., and Smolic, A. (2019).
\newblock Stada: Style transfer as data augmentation.
\newblock {\em arXiv preprint arXiv:1909.01056}.

\end{thebibliography}



\end{document}